\documentclass{SCIS2024}
\usepackage{paralist}
\usepackage{enumitem}
\usepackage{cleveref}
\usepackage{geometry}
\usepackage{arydshln} 
\usepackage{multirow}
\usepackage{multicol}
\usepackage{caption}
\begin{document}
\ArticleType{RESEARCH PAPER}
\Year{2024}
\Month{}
\Vol{}
\No{}
\DOI{}
\ArtNo{}
\ReceiveDate{}
\ReviseDate{}
\AcceptDate{}
\OnlineDate{}

\title{Relative Difficulty Distillation for \\Semantic Segmentation}{Title keyword 5 for citation Title for citation Title for citation}

\author[1, 2]{Dong LIANG}{}
\author[1]{Yue SUN}{}
\author[1]{Yun DU}{}
\author[1]{Songcan CHEN}{{s.chen@nuaa.edu.cn}}
\author[1]{Sheng-Jun HUANG}{}

\AuthorMark{Dong L}

\AuthorCitation{Author A, Author B, Author C, et al}


\address{
\textsuperscript{1}MIIT Key Laboratory of Pattern Analysis and Machine Intelligence, \\
College of Computer Science and Technology, Nanjing University of Aeronautics and Astronautics,\\ Nanjing 211106, China \\
\textsuperscript{2}Nanjing University of Aeronautics and Astronautics Shenzhen Research Institute}

\abstract{
Current knowledge distillation (KD) methods primarily focus on transferring various structured knowledge and designing corresponding optimization goals to encourage the student network to imitate the output of the teacher network. However, introducing too many additional optimization objectives may lead to unstable training, such as gradient conflicts. Moreover, these methods ignored the guidelines of relative learning difficulty between the teacher and student networks. 
Inspired by human cognitive science, in this paper, we redefine knowledge from a new perspective -- the student and teacher networks' relative difficulty of samples, and propose a pixel-level KD paradigm for semantic segmentation named Relative Difficulty Distillation (RDD).
We propose a two-stage RDD framework: Teacher-Full Evaluated RDD (TFE-RDD) and Teacher-Student Evaluated RDD (TSE-RDD). 
RDD allows the teacher network to provide effective guidance on learning focus without additional optimization goals, thus avoiding adjusting learning weights for multiple losses.
Extensive experimental evaluations using a general distillation loss function on popular datasets such as Cityscapes, CamVid, Pascal VOC, and ADE20k demonstrate the effectiveness of RDD against state-of-the-art KD methods. 
Additionally, our research showcases that RDD can integrate with existing KD methods to improve their upper performance bound.

{Codes are available
at \url{https://github.com/sunyueue/RDD.git}.}}

\keywords{knowledge distillation, semantic segmentation, relative difficulty, sample weighting, prediction discrepancy}

\maketitle

\section{Introduction}
Semantic segmentation is the basis of many vision-understanding systems, e.g., medically assisted diagnosis~\cite{1,2,3,4,5}, embodied AI~\cite{6,7,8,9,10}, and driving assistant~\cite{11,12,13,14,15}. 
Current deep neural networks, e.g., DeepLab series~\cite{16,17,18,19}, PSPNet~\cite{20}, HRNet~\cite{21}, have achieved remarkable success. However, these cumbersome models often lead to expensive computation costs. 
To address this issue for embedded and edge deployments, researchers have focused on designing lightweight networks, e.g., ENet~\cite{22}, ICNet~\cite{23}, BiSeNet~\cite{24} and ESPNet~\cite{25}. These networks employ techniques like model quantization~\cite{26} and pruning~\cite{27} to reduce inference cost or utilize knowledge distillation (KD)~\cite{28, 29} to transfer the capabilities of larger models to lightweight ones.

Current KD methods~\cite{30,28,31,29} primarily focus on the design of feature/response-based knowledge and optimization objectives (i.e., distillation loss), aiming to make the student network mimic the teacher's output. 
However, additional optimization objectives accompanied by an imbalance problem may intuitively lead to training instability~\cite{32,33}. 
Furthermore, these approaches often neglect the teacher's ability to delineate learning focus and guide the student network's learning according to the samples' learning difficulty.

In the field of human education and cognition, teachers assess the difficulty of learning materials based on students' needs and adjust the learning difficulty according to their mastery level~\cite{34}. 
This personalized instruction enables students to face appropriate challenges and facilitates the development of higher-order cognitive processes.
From this perspective, we argue that the guidance provided by the teacher network in sample selection and sample difficulty adjustment is equally important for training the student network. 
By incorporating this guidance into knowledge distillation, the accuracy and robustness of student networks can be improved.

\begin{figure}
  \centering
  \subfloat[The formation process of difficulty maps based on confidence for the student and teacher networks]{\includegraphics[width=1.0\linewidth]{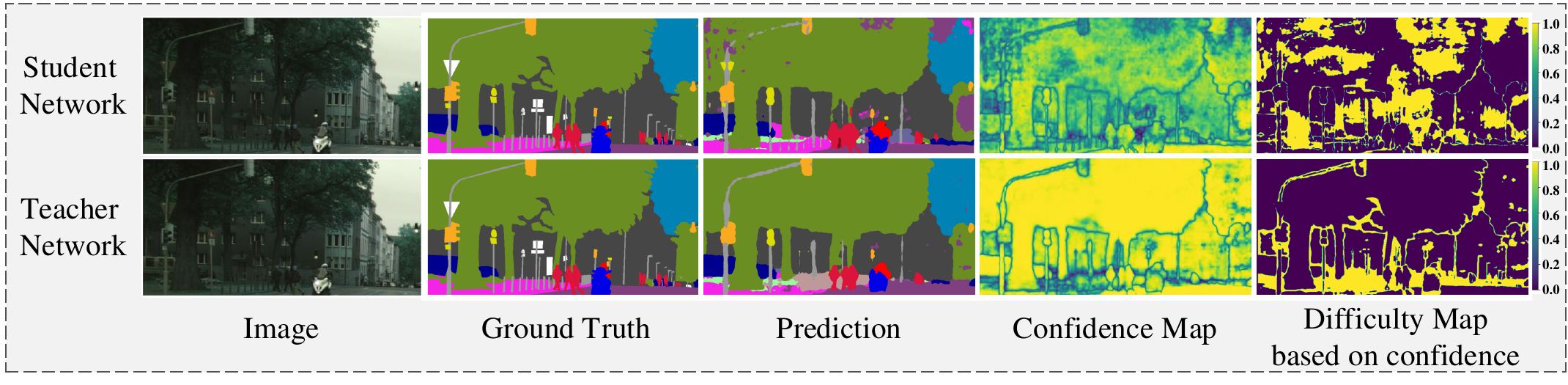}\label{fig:fig1a}}
  \hspace{0.5cm}
  \subfloat[Three different situations of the student and teacher networks evaluating sample difficulty]{\includegraphics[width=1.0\linewidth]{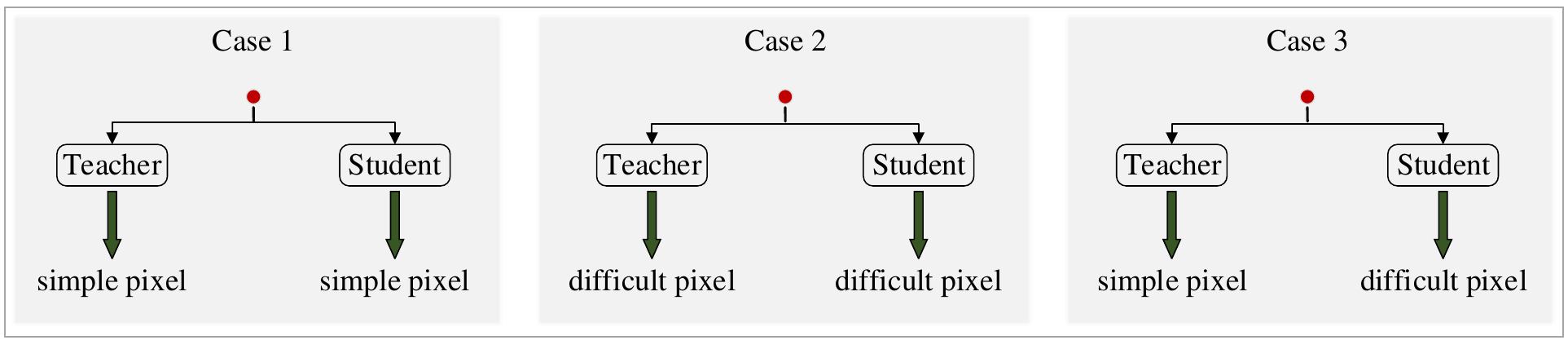}\label{fig:fig1b}}
  
  \caption{The prediction discrepancy based on confidence between the teacher and student networks is used to assess the learning difficulty of pixels.  There are three situations in which the student and teacher networks evaluate the difficulty of a pixel: Case 1, where both networks evaluate the pixel as easy; Case 2, where both networks evaluate the pixel as difficult; Case 3, where there is disagreement on the difficulty of the pixel. Note that for Case 3, the pixel with prediction discrepancies is the difficult pixel that the student network should learn.}
  \label{fig:subfigures}
\end{figure}

Some previous studies~\cite{35,36,37} have evaluated sample difficulty using confidence measures/loss and selectively trained samples based on their difficulty levels.
They consider high-loss/low-confidence samples difficult and low-loss/high-confidence samples easier.
As depicted in Fig.~\ref{fig:fig1a}, for each pixel in an image, the network utilizes the maximum probability of predicted classes as a confidence value to filter difficult pixels, resulting in a confidence map representing the pixel-wise confidence scores.
Subsequently, using a predefined threshold, pixels with confidence values lower than the threshold are marked as 1 (indicating difficult pixels), and those with values higher than the threshold are marked as 0 (indicating simple pixels), thereby generating a difficulty map based on confidence.
The network's performance and generalization capabilities can be enhanced by utilizing this difficulty map to select challenging pixels for training.
However, adding too many difficult pixels can also degrade network performance. \textbf{Therefore, determining which difficult pixels should be mined becomes an essential topic of discussion}.

In Fig.~\ref{fig:fig1b}, we show three situations in which teacher and student networks evaluate the difficulty of a pixel.
The pixel is evaluated to be easy by both the teacher and student network (as shown in Case 1), which is usually the pixel the student network has mastered and no longer needs to spend time on learning.
On the other hand, due to the strait of the semantic segmentation fine-annotation process, the difficult pixel that the teacher network cannot master may even be noisy annotations or have semantic ambiguities. The student network may also struggle to accurately predict the challenging pixel (as shown in Case 2). This class of difficult pixels should also not be learned. 
An ideal valuable difficulty pixel is one that the student network considers difficult and has not fully mastered but is expected to learn well (as shown in Case 3), and the prediction discrepancy based on confidence between the teacher and student networks is used to generate the relative difficulty of pixels can provide such information. 
Therefore, we define knowledge from a new perspective -- the teacher network poses the sample's pixel-level relative learning difficulty as knowledge, providing a guide of divergent samples mining for the learning of the student network.
We also avoid constructing additional optimization objectives to ease instability in network training due to multiple optimization objectives. It also allows integration with other feature/response-based KD methods if desired.

We further discovered that relative difficulty should be in different forms in the learning life cycle.
In the early stage of human education~\cite{38,39}, teachers employ direct instruction to provide students with explicit guidance and a structured learning process, enabling them to quickly acquire fundamental knowledge and skills.  
Similarly, in the early learning stage of knowledge distillation, relying on the student network's judgments to guide training may transmit and amplify errors, making it challenging to correct the student network.
In contrast, the supervisory information from the teacher network is more beneficial for the student network to learn relatively simple pixels and achieve rapid convergence.
On the other hand, adaptive teaching~\cite{40, 41} and personalized teaching~\cite{42} in human education emphasize that teachers should customize the learning process according to learners' current levels and needs to adapt to individual differences and learners' abilities.
Similarly, in knowledge distillation, the student network gains judgment abilities as it progresses through the training stages. 
The student network can assess the sample's difficulty together with the teacher network and adjust its learning accordingly. 
This collaborative approach provides a comprehensive perspective and accurate assessment of sample difficulty, thereby effectively guiding the network to learn relatively difficult pixels in the later learning stage to improve the upper performance bound.

Based on the above discussion, we propose a pixel-level KD paradigm named Relative Difficulty Distillation (RDD) for semantic segmentation. RDD includes two specific distillation methods: teacher-full evaluated RDD (TFE-RDD) and teacher-student evaluated RDD (TSE-RDD), which are two successive learning stages. 
TFE-RDD is designed for the early learning stage when the student network is still far from converging.
It leverages prediction discrepancy between the primary and auxiliary classifiers in the teacher network to acquire relative difficulty knowledge. This knowledge guides the student network to focus on pixels with lower difficulty.
As the learning progresses, TSE-RDD comes into play during the later learning stage. It leverages the prediction discrepancy between the student and teacher networks to generate reliable relative difficulty knowledge. This knowledge progressively guides the student network toward mastering difficult pixels over time.
The proposed RDD scheme effectively distills pixel-level relative difficulty as dark knowledge, guiding the student network to focus more on informative and representative pixels at appropriate learning stages.

The main contributions of this work can be summarized as follows:
\begin{itemize}
\item We propose a new knowledge distillation paradigm for semantic segmentation named Relative Difficulty Distillation (RDD). RDD avoids additional optimization objectives and can seamlessly integrate with other feature/response-based KD methods to improve their upper performance bound.
\item We devise two specific RDD methods, TFE-RDD and TSE-RDD, tailored for the early and later learning stages, respectively. The teacher network incorporates the student network to generate relative difficulty, guiding the student network to focus on the most valuable pixels during the different learning stages.
\item RDD achieves the best distillation performance among the state-of-the-art methods on four popular semantic segmentation datasets with various semantic segmentation architectures.
\end{itemize}




\section{Related Work}\label{sec2}
\subsection{Knowledge distillation for semantic segmentation}\label{subsec2.1}

Knowledge distillation\cite{30} is a method that utilizes the output's probability distribution of the teacher network as the training target of the student network. Methods with similar ideas can be traced back to \cite{43}, which first train an ensemble model and then utilize the reliable output generated by the ensemble model to replace the labels of the original training samples so that the target network can learn. 
In order to deploy deep models on devices with limited resources, Bucilua \emph{et al.}~\cite{44} first proposed model compression technology, which can transfer information from large models or ensemble models to small models for training without significantly reducing accuracy. Subsequently, Hinton \emph{et al.}~\cite{30} summarized and developed this idea and formally proposed the concept of knowledge distillation.

Previous knowledge distillation algorithms~\cite{30,45,46,47,48,49,50,51,52,53} primarily focused on image-level classification tasks. 
However, image-level knowledge distillation seldom considers the locally structured information, so it is congenitally deficient for pixel-level semantic segmentation. 
To alleviate this problem, Liu \emph{et al.}~\cite{28} proposed two structured distillation schemes to transfer structured knowledge from the teacher network to the student network.
Wang \emph{et al.}~\cite{54} introduced intra-class feature variation distillation (IFVD) to transform the teacher network into the student network.  
Shu \emph{et al.}~\cite{31} utilized channel-wise knowledge distillation to minimize the disparity in normalized channel activations between the teacher and the student networks.
Zheng \emph{et al.}~\cite{55} proposed an uncertainty-aware pseudo-label learning approach that leverages uncertainty estimation and knowledge distillation to rectify noisy pseudo labels for domain adaptive semantic segmentation.
Holder \emph{et al.}~\cite{56} presented an efficient uncertainty estimation method for semantic segmentation by transferring uncertainty knowledge from teacher to student networks via distillation.
Ji \emph{et al.}~\cite{57} proposed a contourlet decomposition module (CDM) and a denoised texture intensity equalization module (DTIEM) to mine the structural texture knowledge and enhance the statistical texture knowledge, respectively. 
To establish the global semantic relationships between pixels across different images, Yang \emph{et al.}~\cite{29} attempted to leverage pixel-to-pixel and pixel-to-region relationships as knowledge and transfer global pixel correlation from teachers to students.

The above KD methods mainly transfer various feature/response-based structured knowledge by introducing additional optimization objectives. For example, in ~\cite{57}, the author provided five loss terms to obtain knowledge from low-level and high-level features and force the student network to imitate the teacher network from a broader perspective. Similarly, CIRKD~\cite{29} also provided five loss terms to learn more comprehensive cross-image pixel dependencies.
However, in this multi-objective optimization problem, there are mutual constraints between various objectives, making it difficult for the model to accurately fit the training data, leading to issues such as training instability~\cite{32, 33}. Moreover, more computation and storage time may also be required when computing the structured knowledge of the teacher network, thus slowing down the distillation process. For example, CIRKD~\cite{29} utilized a memory bank for comparative learning, which consumes a certain amount of time when storing pixel embeddings.
In contrast, RDD achieves knowledge transfer by scaling relative gradients without introducing additional optimization objectives, avoiding the trade-off problem between multiple objectives.

\subsection{Training strategies based on curriculum learning}\label{subsec2.2}
Curriculum Learning (CL)~\cite{58} is a training strategy that trains a machine learning model from simple samples to difficult samples by mimicking the meaningful learning sequence in a human curriculum. 
CL can be extended to other methods. 
For example, Self-paced Learning~\cite{59} refers explicitly to a training strategy in which the student network acts as a teacher network and measures the difficulty of training samples based on its losses.
The teacher network is pre-trained in the current dataset or other large-scale datasets, and its knowledge is transferred to the curriculum design of the student network.
Compared with Self-paced Learning~\cite{59}, Transfer Teacher~\cite{60} selects a mature teacher network to evaluate the difficulty of the training samples.
This method addresses the limitation that ``the machine learning network itself may not be mature enough in the initial stage of training to measure sample difficulty accurately."
Although these CL methods have shown effectiveness and ease of use, they often come with increased training costs.
Specifically for semantic segmentation tasks, designing courses or curricula with samples sorted from easy to difficult can lead to additional computational overhead.
Additionally, it has been observed in previous studies~\cite{61} that CL may lead to the loss of correlation between some samples when applied alongside knowledge distillation methods based on sample relationships or graphs. This correlation loss can result in model degradation.

\subsection{Training strategies based on adaptive sample reweighting}\label{subsec2.3}
The concept of reweighting each sample has been extensively studied in the literature.
Some works~\cite{62,63} claim that encouraging models to learn from difficult samples can enhance their upper performance bound. 
For example, Freund \emph{et al.}~\cite{63} selected more difficult samples to train subsequent classifiers.
OHEM~\cite{35}  utilizes the most difficult samples, with the idea that high-loss samples can better train the classifier.
Some studies~\cite{62,64,65} propose soft sampling methods that train the network based on sample importance. 
Focal loss~\cite{62} dynamically assigns higher weights to difficult samples. 
Prime Sample Attention~\cite{64} analyzes evaluation indicators of object detection and assigns weight to positive and negative samples according to different criteria. 
However, Li \emph{et al.}~\cite{65} argued that outliers also exist among difficult instances and introduced the metric of gradient density to adjust the data distribution.
Since the above methods primarily focus on addressing class imbalance problems, they prioritize difficult samples with large training losses and overlook the contribution of simple samples.

In this paper, we similarly incorporate the idea of sample reweighting to generate reliable relative difficulty knowledge by exploiting prediction disagreement between student and teacher networks.
Weighting the samples based on generated relative difficulty knowledge can guide the student network to focus its learning on valuable easy or difficult pixels at different stages of training. 
\\

\section{Methodology}\label{sec3}
In this section, we show RDD-based semantic segmentation through the relative difficulty knowledge $RD$ provided by the teacher and student networks.
We begin by introducing the loss function paradigm for the semantic segmentation with knowledge distillation in Subsection \ref{subsec3.1}. 
Next, we present the generalized definition of the relative difficulty knowledge $RD$ in Subsection \ref{subsec3.2}.
Subsequently, we defined the two-stage distillation based on relative difficulty and the corresponding generated knowledge $\mathit{RD}_\mathit{TFE}$ and $\mathit{RD}_\mathit{TSE}$ in subsection \ref{subsec3.3}, respectively.
The implementation of Teacher-Full Evaluated RDD (TFE-RDD) and Teacher-Student Evaluated RDD (TSE-RDD) are in Subsection \ref{subsec3.4} and \ref{subsec3.5}.

\subsection{Preliminary}\label{subsec3.1}

 \textbf{Loss function paradigm for semantic segmentation.} 
Semantic segmentation classifies each pixel in an image from $C$ categories to corresponding category labels.
The segmentation network typically takes an RGB image with dimensions $W\times H\times 3$ as input. It extracts the dense image feature map $F$ using a backbone architecture, where $H$ and $W$ represent the height and width of the image. The categorical logit map \textbf{Z} is generated from feature map $F$ using the classifier by applying a classifier. 
Optimization is then performed using a cross-entropy loss:
 
\begin{equation}
\begin{aligned}
        L_{task}=\frac{1}{H \times W} \sum_{h=1}^{H} \sum_{w=1}^{W} C E\left(\sigma\left(\mathbf{Z}_{h, w}\right), \mathbf{y}_{h, w}\right),
	\label{eq:loss_task}
\end{aligned}
\end{equation}
where $\mathbf{y}_{h,w}$ denotes the ground-truth label for the ${(h,w)}$-th pixel, and \textbf{Z}$_{h, w}$ denotes the output logits for the ${(h,w)}$-th pixel. 
The softmax function $\sigma$ generates the category probability, and $CE$ is the cross-entropy loss to measure the difference between the ground truth and category probability.
\\
\textbf{Loss function paradigm for pixel-wise knowledge distillation.} 
Knowledge distillation in semantic segmentation tasks generally employs a point-wise alignment to learn structural knowledge among spatial locations. It can be formulated as follows:
\begin{equation}
    \begin{aligned}
    {
    L_{kd}=\frac{1}{H \times W} \sum_{h=1}^{H} \sum_{w=1}^{W} K L\left[\sigma\left(\frac{\mathbf{Z}_{h, w}^{s}}{T}\right) \| \sigma\left(\frac{\mathbf{Z}_{h, w}^{t}}{T}\right)\right],
        }
    \end{aligned}
	\label{eq:loss_kd}
\end{equation}
\begin{figure}[ht]
    \centering
    \includegraphics[width=1.0\textwidth]{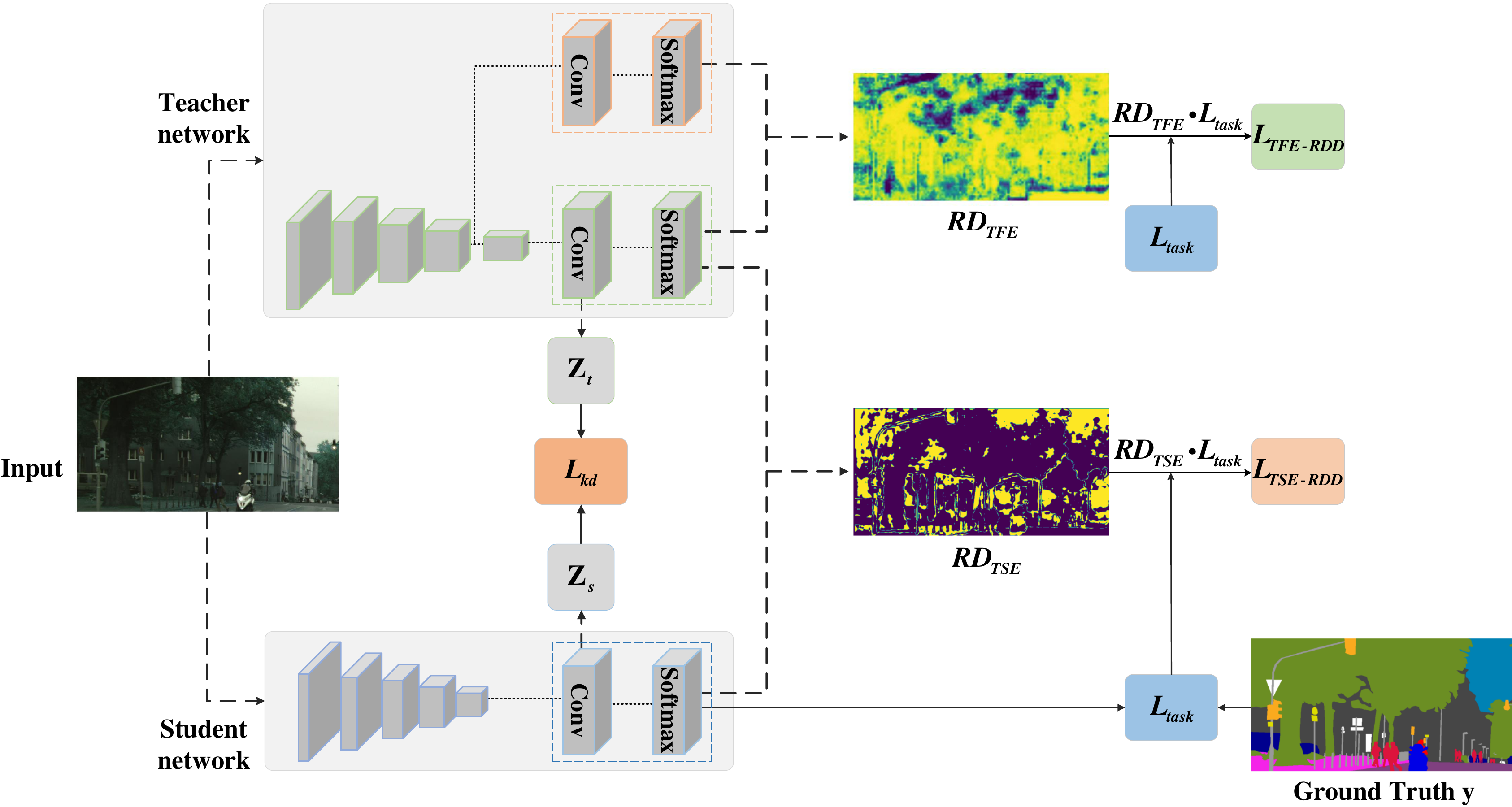} 
    \caption{The proposed Relative Difficulty Distillation (RDD).
    }
    \label{Fig:rdd}
\end{figure}
where \textbf{Z}$_{h,w}^{s}$ and \textbf{Z}$_{h,w}^{t}$ represent the output logits for the ${(h,w)}$-th pixel produced from the student and the teacher network, respectively. 
$\sigma$ function calculates the category probability of the $(h,w)$-th pixel generated by the student and teacher networks, respectively.
$KL$ denotes the Kullback-Leibler divergence, which measures the difference between two probability distributions. The parameter $T$ represents the temperature taken by distillation and reflects the label's softening degree. 
For a fair comparison with previous works~\cite{28, 29}, we set $T$ =1 in our experiments.

\subsection{Define relative difficulty as knowledge}\label{subsec3.2}
A well-trained teacher network can provide the student network with relative difficulty knowledge, denoted as $RD$ $\in \mathbb{R}^{H \times W}$, where $\mathbb{R}$ represents the Euclidean space.
We use $RD$ to measure the difficulty of samples. The value of $RD$ indicates the level of difficulty associated with each sample. Larger values correspond to more difficult samples. 
We apply the proposed $RD$ to the task loss $L_{task}$ calculation. 
The loss $L_{task}$ with relative difficulty knowledge $RD$ can be formalized as $w(L_{task}, RD)$. $w$ imposes sample weights on $L_{task}$ based on the $RD$.
Intuitively, $RD$ increases the loss contribution related to the valuable pixel sample in the current stage without introducing additional training objectives.
The final KD loss is formulated as
\begin{equation}
\begin{aligned}
       L_{RDD} = w(L_{task}, RD) + L_{kd},
	\label{eq:loss_task1} 
\end{aligned}
\end{equation}
where $w$ leverages $RD$ to guide learning during the calculation of $L_{task}$, and $L_{kd}$ provides additional supervision.
The $w$ function and $L_{kd}$ loss represent concrete implementations of difficulty-based distillation and feature/response-based distillation, respectively. These two distillation schemes are independent and can be integrated additively.
In our proposed method, the cross-entropy loss is used for calculating $L_{task}$, and $L_{kd}$ refers to the distillation loss mentioned in Section \ref{subsec3.1}. 

\subsection{Define two-stage relative difficulty distillation}\label{subsec3.3}

During the initial stages of training, the student network lacks the ability to accurately predict pixels and requires full guidance from the teacher network to determine the learning difficulty. The student network achieves fast convergence by focusing on relatively simple pixels.
As dynamic learning progresses, the student network gradually becomes capable of accurately fitting most simple pixels and develops its own judgment capabilities. 
At this stage, the teacher network needs to select appropriate, challenging pixels based on the student network's needs and skill level to promote the development of the student network's capabilities.
Consequently, we believe $RD$ should be diversified and progressive.
Based on the above inspiration, we propose two stages of RDD: Teacher-Full Evaluated RDD (TFE-RDD) in the early learning stage and Teacher-Student Evaluated RDD (TSE-RDD) in the later learning stage. TFE-RDD and TSE-RDD provide two types of relative difficulty knowledge: \textbf{Teacher-Full Evaluated Relative Difficulty} (denoted as $\mathit{RD}_\mathit{TFE}$) and \textbf{Teacher-Student Evaluated Relative Difficulty} (denoted as $\mathit{RD}_\mathit{TSE}$) respectively.

As depicted in Fig.~\ref{Fig:rdd}, in the TFE-RDD stage, RDD relies on the teacher network to provide full guidance on the learning difficulty. 
The prediction discrepancy between the teacher network's primary and auxiliary classifiers can provide $\mathit{RD}_\mathit{TFE}$ to the student network, which uses $\mathit{RD}_\mathit{TFE}$ to focus training on simple pixels.
We incorporate $\mathit{RD}_\mathit{TFE}$ into $L_{task}$ calculation as follows: 

\begin{equation}
\begin{aligned}
       L_{TFE-RDD} = \mathit{RD_{TFE}} \cdot L_{task}.
	\label{eq:loss_task2} 
\end{aligned}
\end{equation}
As the student network matures, it becomes necessary for the teacher network to consider the student's relative difficulty in teaching according to their disagreement.
In the TSE-RDD stage, we believe that leveraging the discrepancy between the teacher and student networks' prediction confidences will generate $\mathit{RD}_\mathit{TSE}$. The student network utilizes $\mathit{RD}_\mathit{TSE}$ to focus training on valuable difficult pixels.
We apply $\mathit{RD}_\mathit{TSE}$ to impose learning attention during $L_{task}$ calculation as follows: 

\begin{equation}
\begin{aligned}
       L_{TSE-RDD} = \mathit{RD_{TSE}} \cdot L_{task}.
	\label{eq:loss_task3} 
\end{aligned}
\end{equation}
The following sections provide further details about TFE-RDD and TSE-RDD separately.

\begin{figure}[!t]
    \centering
    \includegraphics[width=1.0\textwidth]{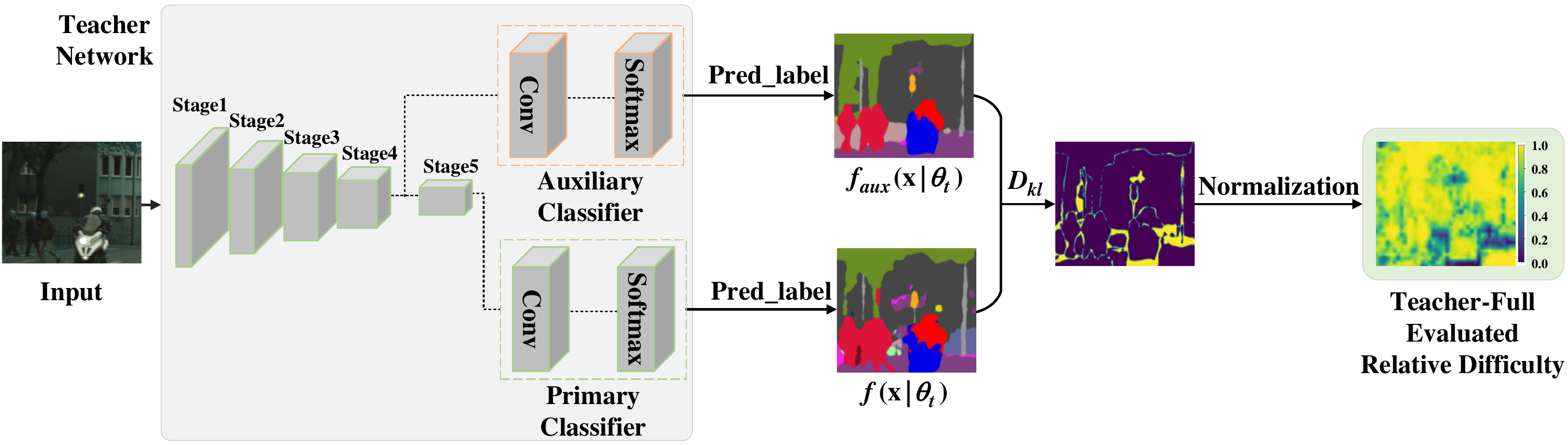} 
    \caption{In the TFE-RDD stage, the difficulty map based on prediction discrepancy is obtained using the prediction results of primary and auxiliary classifiers of the teacher network, and the student network is guided to learn simple pixels for efficient fitting.}
    \label{Fig:TERDD}
\end{figure}

\subsection{Teacher-full evaluated RDD (TFE-RDD)}\label{subsec3.4}

Most semantic segmentation networks incorporate double classification heads (primary and auxiliary classifier) to alleviate the gradient disappearance issue, including~\cite{20, 16, 17} and the modified DeepLab-v2 in~\cite{66, 67, 68}.
Previous works~\cite{69,70} have shown that a difference in predictions between the \textbf{primary} and \textbf{auxiliary} classifiers indicates the sample is relatively difficult to classify.

In the left part of Fig.~\ref{Fig:TERDD}, we present a simplified schematic diagram of a dual classifier network based on DeepLabV3~\cite{17} with ResNet-101~\cite{71} as its backbone. 
The ResNet-101 network consists of five stage blocks, which produce five feature maps capturing different scales and semantic information.
We introduce an auxiliary classifier with an identical structure to the primary classifier, comprising a convolutional kernel and softmax function. 
The primary classifier employs the feature map acquired from stage 5 as input, while the auxiliary classifier uses the feature map obtained from stage 4. 
Since each stage's feature map corresponds to distinct semantic information and context, both classifiers classify the input based on features from different stages. 
Any differences in these features will result in different predictions. 
Difficult pixels often exhibit more significant differences in feature maps across different stages, as they have more complex feature representations. Consequently, the two classifiers may produce inconsistent predictions for these challenging pixels.

In the early stage of training, the student network has not yet converged, and the prediction discrepancy between the student network's double classifiers is not a criterion for determining whether the sample is difficult to classify. 
Therefore, the pre-trained teacher network is more suitable for guiding the student network in the early learning stage.
We leverage the prediction discrepancy between the teacher network's primary and auxiliary classifiers to reflect the difficulty of pixels.

We first utilize the teacher network's primary classifier $f\left(\cdot \mid \theta_{t}\right)$ and auxiliary classifier $f_{aux}\left(\cdot \mid \theta_{t}\right)$ to predict the label of each pixel.
The KL divergence ($D_{k l}$) is then utilized to calculate the discrepancy between the two predicted labels:
\begin{figure}[!t]
    \centering
    \includegraphics[width=1.0\textwidth]{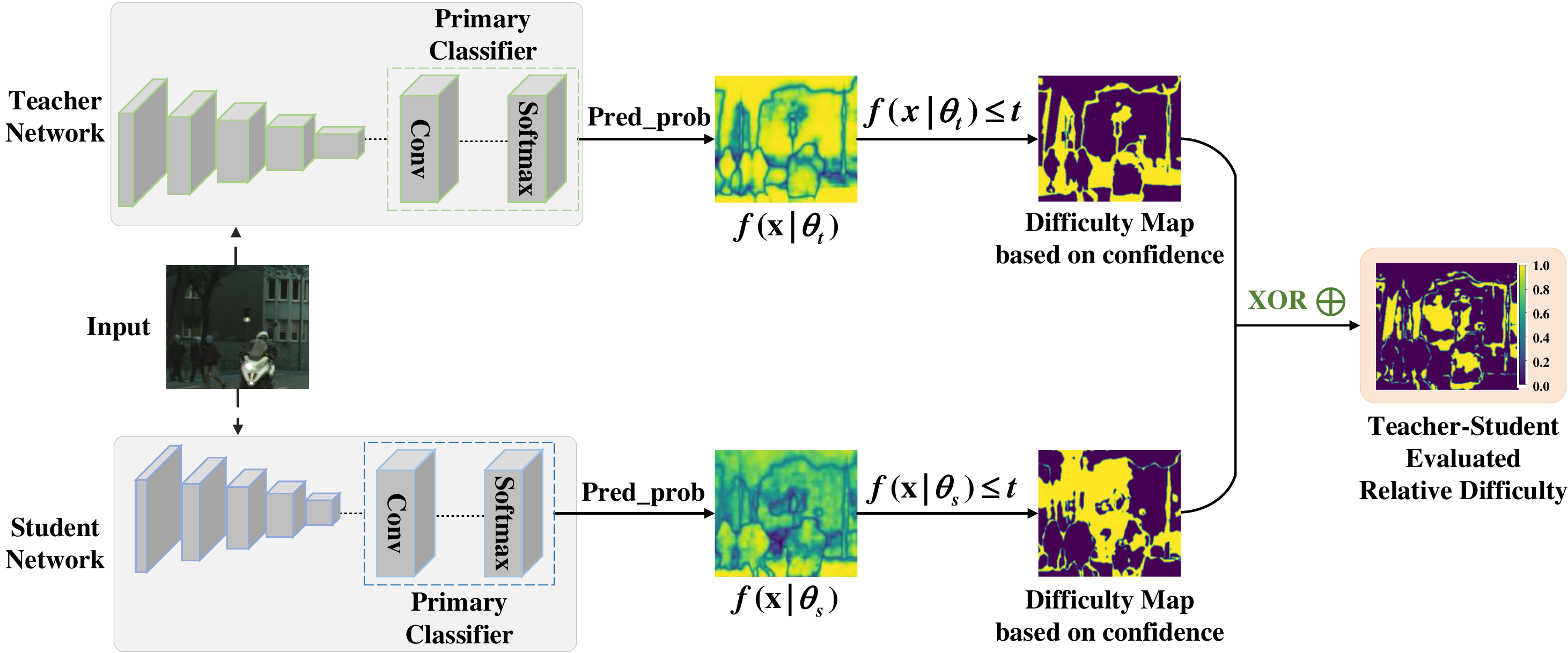} 
    \caption{In the TSE-RDD stage, the difficulty maps based on confidence are obtained using the prediction results of the teacher and student networks. The filtered difficulty maps are applied to Exclusive-OR operations to obtain valuable difficult pixels and expand the upper performance bound.}
    \label{Fig:TSRDD}
\end{figure}

\begin{equation}
\begin{aligned}
        D_{k l}=f\left(x \mid \theta_{t}\right) \log \left(\frac{f\left(x \mid \theta_{t}\right)}{f_{a u x}\left(x \mid \theta_{t}\right)}\right),
\end{aligned}
	\label{eq:d_kl}
\end{equation}
where $\theta_{t}$ represents the parameter set of teacher network. After obtaining the $D_{k l}$ for each pixel via KL divergence calculation, we normalize it within a $[0,1]$ range to obtain the pixel's relative difficulty values. These values are referred to as Teacher-Full Evaluated Relative Difficulty ($\mathit{RD}_\mathit{TFE}$), which is calculated as follows:

\begin{equation}
\begin{aligned}
        \mathit{RD_{TFE}}=\exp \left\{-D_{k l}\right\}.
	\label{eq:TERD}
\end{aligned}
\end{equation}
The value of $\mathit{RD}_\mathit{TFE}$ reflects how difficult each pixel is classified, and larger values indicate simpler pixels. 
Subsequently, we apply $\mathit{RD}_\mathit{TFE}$ to the semantic segmentation task loss ($L_{task}$) to get the total loss for the current stage using the following formula:
\begin{equation}
    L_{RDD} = L_{TFE-RDD} +L_{kd} = \mathit{RD_{TFE}} \cdot L_{task} +L_{kd}.
	\label{eq:loss_TERDD}
\end{equation}
By adjusting the loss weight for each pixel based on $\mathit{RD}_\mathit{TFE}$, simple pixels have larger values of $\mathit{RD}_\mathit{TFE}$ for the student network to learn in this stage.  
\subsection{Teacher-student evaluated relative difficulty distillation (TSE-RDD)}\label{subsec3.5}
As the student network gradually converges, difficult pixels become more helpful in further improving the network's performance.
However, it is essential to note that adding too many difficult pixels can adversely affect the student network's performance.
According to observations from previous works~\cite{72, 73}, deep learning networks often suffer from overconfidence and tend to overfit difficult pixels.
Additionally, after the initial training stage, the student network develops an ability to judge the difficulty of samples to some extent. Relying solely on the relative difficulty provided by the teacher network does not consider the student network's needs.
Hence, it is crucial for the teacher network to assess the sample's learning difficulty based on the student network's requirements and adjust the learning difficulty according to their disagreements.

Considering the aforementioned factors, we propose a method that utilizes the relative difficulty generated by the discrepancy based on prediction confidence between the teacher and student networks, as depicted in Fig.~\ref{Fig:TSRDD}.
First, the teacher and student networks discard pixels with confidence values greater than a threshold $t$ while retaining those difficult pixels that still need to be learned. This filtering process generates the corresponding difficulty maps based on confidence.
We then perform an \rm{XOR} operation between the filtered difficulty maps of both networks to obtain a relative difficulty map named $\mathit{RD}_\mathit{TSE}$. The formulation for computing $\mathit{RD}_\mathit{TSE}$ is as follows:

\begin{equation}
    \mathit{RD_{TSE}} = (f(x \mid \theta_{s}) \leq t) \oplus (f(x \mid \theta_{t}) \leq t).
	\label{eq:TSRD}
\end{equation}
$\mathit{RD}_\mathit{TSE}$ takes a value of 0 or 1. 
A value of 0 indicates two cases: 
\begin{inparaenum}[1)]
  \item The current pixel is simple and can be easily mastered by both the teacher and student networks. Such pixels can be unlearned at the current stage.
  \item The current pixel is difficult for both networks to master. Difficult pixels that even the teacher network cannot master may contain noisy annotations or semantic ambiguities, so there is no need for the student network to learn them.
\end{inparaenum}
On the other hand, a value of 1 indicates that the current pixel is the difficult pixel that the teacher has mastered while the student has not. This part of the pixels is the one that the student network is expected to learn well.
The method of selecting pixels at the current stage is analogous to the teacher's highlighting operation in human education: most students usually cannot surpass the teacher's level, and if they expect to reach the teacher's level, then they only need to enhance their learning of the parts of the content that the student has not learned well, but the teacher has learned well.

After that, we apply $\mathit{RD}_\mathit{TSE}$ to $L_{task}$ to get the total loss of the current stage, which is calculated as
\begin{equation}
    L_{RDD} = L_{TSE-RDD} +L_{kd} = \mathit{RD_{TSE}} \cdot L_{task} +L_{kd}.
	\label{eq:loss_TSRDD}
\end{equation}

By adjusting the loss weight of each pixel using $\mathit{RD}_\mathit{TSE}$, valuable difficult-to-classify pixels (with a corresponding $\mathit{RD}_\mathit{TSE}$ value of 1) receive more attention during training.

\begin{algorithm}
\caption{Relative Difficulty Distillation}\label{alg:algorithm}
\textbf{Input}: Input images $x$, labels $y$, the parameter of teacher network $\theta_{t}$, the iterations of TFE-RDD $iter_{TFE-RDD}$, cross-entropy loss function $CE$, Kullback-Leibler divergence function $KL$, Softmax function $\sigma$. 
\textbf{Output}: The parameter of student network $\theta_{s}$.
\begin{algorithmic}[1]
    \WHILE{the student network has not converged}
    \STATE $L_{task} = \frac{1}{H \times W} \sum\limits_{h=1}^{H} \sum\limits_{w=1}^{W} C E\left(\sigma\left(\mathbf{Z}_{h, w}\right), \mathbf{y}_{h, w}\right)$ 
    \STATE $L_{kd}=\frac{1}{H \times W} \sum\limits_{h=1}^{H} \sum\limits_{w=1}^{W} K L\left(\sigma\left(\frac{\mathbf{Z}_{h, w}^{s}}{T}\right) \| \sigma\left(\frac{\mathbf{Z}_{h, w}^{t}}{T}\right)\right)$
    \IF{$iter \leq iter_{TFE-RDD}$ }
        \STATE $D_{k l}=f\left(x \mid \theta_{t}\right) \log \left(\frac{f\left(x \mid \theta_{t}\right)}{f_{a u x}\left(x \mid \theta_{t}\right)}\right)$
        \STATE $\mathit{RD_{TFE}}=\exp \left\{-D_{k l}\right\}$
        \STATE $L_{RDD}=\mathit{RD_{TFE}} \cdot L_{task}  + L_{kd}$
    \ELSE
        \STATE $\mathit{RD_{\mathit{TSE}}} = (f(x \mid \theta_{s}) \leq t) \oplus (f(x \mid \theta_{t}) \leq t)$
        \STATE $L_{RDD} = \mathit{RD_{TSE}} \cdot L_{task} + L_{kd}$
    \ENDIF
    \STATE $L_{overall}.backward()$
        
    \ENDWHILE
    \STATE \textbf{return} $\theta_s$
\end{algorithmic}
\end{algorithm}

Algorithm~\ref{alg:algorithm} provides the pseudo-code illustrating the overall training pipeline of RDD. The proposed RDD learning paradigm leverages the teacher and student networks to deliver the relative difficulty of pixels to guide the network's learning.

\subsection{Integrating with other approaches}\label{subsec3.6}
Since RDD only affects the $L_{task}$ loss, it can be seamlessly integrated with other methods without introducing additional optimization objectives. In this subsection, we demonstrate how to incorporate RDD into AT~\cite{52}, DSD~\cite{74}, and CIRKD~\cite{29} with multiple distillation losses, deriving the corresponding distillation loss formulations.
To ensure consistency with the comparison results of other experiments and to conveniently integrate with other methods, we keep the hyperparameters in the integrated loss function consistent with those of the comparison method. Specific hyperparameter values are provided in each subsection below.

\subsubsection{The distillation loss of AT method after integrating RDD}\label{subsec3.6.1}

The total loss defined by AT~\cite{52}:
\begin{equation}
L_{AT}=L_{task}+\frac{\beta}{2}\sum_{j\epsilon I}^{}||\frac{{Q}_{ S}^{j}}{||{Q}_{ S}^{j}||_2}-\frac{{Q}_{ T}^{j}}{||{Q}_{T}^{j}||_2} ||_2,
\end{equation}
where ${{Q}_{S}^{j}}$ and ${{Q}_{T}^{j}}$ are respectively the $j$-th pair of student and teacher attention maps in vectorized form, $\frac{{Q}_{ S}^{j}}{||{Q}_{ S}^{j}||_2}$ and $\frac{{Q}_{ T}^{j}}{||{Q}_{T}^{j}||_2}$ are the result of using $l_{2}$-normalization attention maps. 
The calculation details of $\frac{\beta}{2}\sum\limits_{j\epsilon I}^{ }||\frac{\mathbf{Q}_{ S}^{j}}{||\mathbf{Q}_{ S}^{j}||_2}-\frac{\mathbf{Q}_{ T}^{j}}{||\mathbf{Q}_{T}^{j}||_2} ||_2$ are described in AT~\cite{52}. $L_{task}$ is the semantic segmentation task loss function mentioned in Section \ref{subsec3.1}, represented by the cross entropy function. 

The distillation loss of RDD :
\begin{align}L_{RDD}=\left\{\begin{aligned}
    \mathit{RD_{TFE}} \cdot L_{task} + L_{kd},\quad iter \leq iter_{TFE-RDD},\\
    \mathit{RD_{TSE}} \cdot L_{task}+L_{kd},\quad iter > iter_{TFE-RDD},
\end{aligned}\right.\end{align}
where $\mathit{RD}_\mathit{TFE}$ represents the relative difficulty generated by the teacher network, and $\mathit{RD}_\mathit{TSE}$ denotes the relative difficulty generated by the student-teacher cooperation. The calculations of $\mathit{RD}_\mathit{TFE}$ and $\mathit{RD}_\mathit{TSE}$ are described in Algorithm~\ref{alg:algorithm} in our paper. $iter_{TFE-RDD}$ is the iteration number of the TFE-RDD stage. 
$L_{kd}$ is the pixel-level distillation loss mentioned in Section \ref{subsec3.1}, represented by Kullback-Leibler (KL) divergence. 

The distillation loss of AT~\cite{52} method after integrating RDD can be derived:

\begin{align}L_{AT\_RDD}=\left\{\begin{aligned}
\mathit{RD_{TFE}} \cdot L_{task}+\frac{\beta}{2}\sum_{j\epsilon I}^{}||\frac{{Q}_{ S}^{j}}{||{Q}_{ S}^{j}||_2}-\frac{{Q}_{ T}^{j}}{||{Q}_{T}^{j}||_2} ||_2   ,\quad iter \leq iter_{TFE-RDD},\\
   \mathit{RD_{TSE}}\cdot L_{task}+\frac{\beta}{2}\sum_{j\epsilon I}^{}||\frac{{Q}_{ S}^{j}}{||{Q}_{ S}^{j}||_2}-\frac{{Q}_{ T}^{j}}{||{Q}_{T}^{j}||_2} ||_2,\quad iter > iter_{TFE-RDD}.
\end{aligned}\right.\end{align}

In the new loss term described above,  we follow the default parameter settings in AT~\cite{52} and set the weighting parameter $\beta$ to $10^3$ divided by the number of elements in the attention map and batch size for each layer.
We only modify the weights for each pixel by incorporating the relative difficulty factors obtained through RDD in front of the $L_{task}$.

\subsubsection{The distillation loss of DSD method after integrating RDD }\label{subsec3.6.2}

The total loss defined by DSD~\cite{74}:
\begin{equation}
    L_{DSD}=L_{task}+\alpha L_{PSD}+\beta L_{CSD},
\end{equation}
where $L_{PSD}$ refers to the pixel-wise similarity distillation loss between the teacher and student networks, $L_{CSD}$ refers to the category-wise similarity distillation loss between the teacher and student networks, and $\alpha$ and $\beta$ are the weight balance parameters. The calculation details of $L_{PSD}$ and $L_{CSD}$ are described in DSD~\cite{74}. 

The distillation loss of RDD :
\begin{align}L_{RDD}=\left\{\begin{aligned}
    \mathit{RD_{TFE}} \cdot L_{task} + L_{kd},\quad iter \leq iter_{TFE-RDD},\\
    \mathit{RD_{TSE}} \cdot L_{task}+L_{kd},\quad iter > iter_{TFE-RDD}.
\end{aligned}\right.\end{align}

Hence, we derive the modified distillation loss for the DSD method after integrating RDD:
\begin{align}L_{DSD\_RDD}=\left\{\begin{aligned}
\mathit{RD_{TFE}} \cdot L_{task}+\alpha L_{PSD}+\beta L_{CSD},\quad iter \leq iter_{TFE-RDD},\\
    \mathit{RD_{TSE}}\cdot L_{task}+\alpha L_{PSD}+\beta L_{CSD},\quad iter > iter_{TFE-RDD}.
\end{aligned}\right.\end{align}

In the new loss term described above, we follow the default parameter settings in DSD~\cite{74}, setting the weighting parameter $\alpha$ to $10^3$ and $\beta$ to 10.
It allows the student network to prioritize learning pixels that hold greater value at a given stage. Introducing this bias towards valuable pixel learning enhances the student network's overall performance.

\subsubsection{The distillation loss of CIRKD method after integrating RDD }\label{subsec3.6.3}

The total loss defined by CIRKD~\cite{29}:
\begin{equation}
     L_{CIRKD}=L_{task}+L_{kd}+\alpha L_{batch\_p2p}+\beta L_{memory\_p2p}+\gamma L_{memory\_p2r},
\end{equation}
where $L_{batch\_p2p}$ represents distillation loss of mini-batch-based pixel-to-pixel, $L_{memory\_p2p}$ denotes distillation loss of memory-based pixel-to-pixel, $L_{memory\_p2r}$ denotes distillation loss of memory-based pixel-to-region. $\alpha$, $\beta$ and $\gamma$ are the weight balance parameters. 
The further calculation details of $L_{batch\_p2p}$, $L_{memory\_p2p}$ and $L_{memory\_p2r}$ are described in~\cite{29}.

The distillation loss of RDD :
\begin{align}L_{RDD}=\left\{\begin{aligned}
    \mathit{RD_{TFE}} \cdot L_{task} + L_{kd},\quad iter \leq iter_{TFE-RDD},\\
    \mathit{RD_{TSE}} \cdot L_{task}+L_{kd},\quad iter > iter_{TFE-RDD}.
\end{aligned}\right.\end{align}

After integrating RDD, the modified distillation loss of the CIRKD method is derived as follows:

\begin{align}L_{CIRKD\_RDD}=\left\{\begin{aligned}
\mathit{RD_{TFE}}\cdot L_{task}+L_{kd}+\alpha L_{batch\_p2p}+\beta L_{memory\_p2p}+\gamma L_{memory\_p2r},\\iter \leq iter_{TFE-RDD},\\
\mathit{RD_{TSE}}\cdot L_{task}+L_{kd}+\alpha L_{batch\_p2p}+\beta L_{memory\_p2p}+\gamma L_{memory\_p2r},\\iter > iter_{TFE-RDD}.
\end{aligned}\right.\end{align}

In this new loss term, we follow the default parameter settings in CIRKD~\cite{29}, setting the weighting parameter $\alpha$ to 1, $\beta$ to 0.1, and $\gamma$ to 0.1.
Moreover, RDD can seamlessly integrate with other semantic segmentation methods based on knowledge distillation. This integration allows us to enhance the performance of the student network further from existing approaches.

\section{Experiments}\label{sec4}
\subsection{Data descriptions}\label{subsec4.1}
We evaluate our method on the Cityscapes~\cite{75}, CamVid~\cite{76}, PASCAL VOC 2012~\cite{77} and ADE20k~\cite{78}. 
Cityscapes~\cite{75} is a large-scale dataset for urban street scenes with 5000 high-quality images with pixel-wise annotations with 19 semantic classes. 
These finely annotated images are divided into 2975/500/1525 for train/val/test. 
CamVid~\cite{76} is an automotive driving dataset that contains 367/101/233 images for train/val/test with 11 semantic classes. 
Pascal VOC 2012~\cite{77} is a general object segmentation benchmark with 21 classes. 
The original segmentation dataset is divided into 10582/1449/1456 for train/val/test. 
Following previous works, we also use the extra annotations provided by~\cite{79}. ADE20K~\cite{78} is a scene segmentation dataset covering 150 fine-grained semantic classes with 20210 images. 

\subsection{Implementation details}\label{subsec4.2}
\textbf{Network architectures}.
We employ DeepLabV3~\cite{17} with a ResNet-101 backbone~\cite{71} as the cumbersome teacher network for all experiments. As for student networks, we utilize various segmentation architectures to validate the effectiveness of our distillation methods. Specifically, we use DeepLabV3~\cite{17} and PSPNet~\cite{20} with ResNet-18 backbone and MobileNetV2~\cite{80} backbone for the student network.
\\
\textbf{Training details}. The networks are trained using mini-batch stochastic gradient descent (SGD) with a momentum of 0.9 and weight decay of 0.0005.
For Cityscapes, CamVid, and PASCAL VOC 2012 (VOC12), we set the number of iterations to 40,000; for the ADE20k dataset, we set it to 80,000.
The learning rate is initialized at 0.02 and is multiplied by $(1 - \frac{iter}{ iter_{total}}) ^ {0.9}$ during training. We randomly crop images into sizes of 512 ×1024, 360 ×360, and 512 ×512 for Cityscapes, CamVid, Pascal VOC, and ADE20k datasets, respectively. Normal data augmentation techniques such as random flipping and scaling in the range of [0.5, 2] are applied during training.  
The threshold $t$ of TSE-RDD in Eq.~(\ref{eq:TSRD}) is set as 0.7 (to be evaluated in the ablation study). The temperature $T$ in $L_{kd}$ is set to be 1. 
All experiments are conducted on four 3090 GPUs using mixed-precision training.
\\
\textbf{Evaluation metrics}. Following the standard setting, we adopt the mIoU (mean intersection over union) metric to evaluate the performance of different methods. mIoU calculates the ratio of the intersection and union of two sets of true and predicted labels.

\subsection{Comparison with existing methods on four datasets}\label{subsec4.3}

In this section, we compare our proposed RDD with recent knowledge distillation-based semantic segmentation methods, including SKD~\cite{28}, IFVD~\cite{54}, CWD~\cite{31}, CIRKD~\cite{29}, AT~\cite{52} and DSD~\cite{74} on four representative semantic segmentation datasets described above.
The experimental results are shown in ~\cref{table:results_cityscapes,table:results_voc2012,table:results_ade20k,table:results_camvid}.
In the table,`` T: DeepLabV3-Res101" denotes training with the teacher network. `` S: DeepLabV3-Res18", `` S: DeepLabV3-Res18*", `` S: DeepLabV3-MBV2", and `` S: PSPNet-Res18" denote training with the student networks.
It should be noted that we used DeepLabV3-MBV2 as the student network, which uses the lightweight MobileNetV2 as the backbone network. This setup aims to validate the performance of RDD on lightweight networks in order to explore the possibility of deploying our approach on mobile devices.


\begin{table}[!h]
\footnotesize
\caption{Performance comparison with SOTA KD methods over various student segmentation networks on Cityscapes. FLOPs are measured based on the test size of 1024 × 2048. * denotes that we do not initialize the backbone with ImageNet~\cite{81} pre-trained weights.}
\label{table:results_cityscapes}
\begin{tabular*}{1.0\textwidth}{@{\extracolsep\fill}>{\centering\hspace{1em}}c>{\centering\hspace{5em}}c>{\centering\hspace{3em}}cc}
\toprule[1.5pt]
 \hspace*{1cm}
        Method & \centering mIoU $(\%)$ & FLOPs (G) &  Params (M)  \hspace*{1cm}\\ 
        \midrule[0.8pt]
        
        \multicolumn{4}{c}{Some related non-KD semantic segmentation methods} \\
        \midrule[0.8pt]
        ENet~\cite{22} & 58.42 & 14.4G &   0.35M \hspace*{1cm}\\ 
        ESPNet~\cite{25} & 60.34 & 17.7G &   0.36M \hspace*{1cm}\\
        ICNet~\cite{23} & 69.53 & 113.2G &   26.5M \hspace*{1cm}\\  
        FCN~\cite{82} & 62.76 & 1335.6G &   134.5M \hspace*{1cm}\\ 
        RefineNet~\cite{83} & 73.58 & 2102.8G &   118.1M \hspace*{1cm}\\
        OCNet~\cite{84} & 80.12 & 2194.2G &   62.6M \hspace*{1cm}\\
        \midrule[1.0pt]
        \multicolumn{4}{c}{Comparison with different KD methods} \\
        
        \midrule[1.0pt]
		T: DeepLabV3-Res101 & 78.07 & 2371G &   61.1M \hspace*{1cm}\\ 
        \midrule[0.8pt]
         \hspace*{1cm}
		S: DeepLabV3-Res18 & 74.21 & \multirow{6}{*}{572.0G} & \multirow{6}{*}{13.6M}  \hspace*{1cm}\\ 
   \hspace*{1cm}
		+ SKD~\cite{28} & 75.42 $(\uparrow1.21)$ & &   \hspace*{1cm}\\
   \hspace*{1cm}
		+ IFVD~\cite{54} & 75.59 $(\uparrow1.38)$ & &   \hspace*{1cm}\\
   \hspace*{1cm}
		+ CWD~\cite{31} & 75.55 $(\uparrow1.34)$ & &   \hspace*{1cm}\\
   \hspace*{1cm}
		+ CIRKD~\cite{29} & 76.38 $(\uparrow2.17)$ & &   \hspace*{1cm}\\ 
   \hspace*{1cm}
		+ RDD (Ours) & \textbf{77.18 (\bm{$\uparrow\mkern-5mu$}2.97)} & &   \hspace*{1cm}\\ 
        \hdashline[2pt/1pt]
         \hspace*{1cm}
		S: DeepLabV3-Res18* & 65.17  & \multirow{6}{*}{572.0G}  &  \multirow{6}{*}{13.6M} \hspace*{1cm}\\
   \hspace*{1cm}
		+ SKD~\cite{28} & 67.08 $(\uparrow1.91)$ &  &   \hspace*{1cm}\\ 
   \hspace*{1cm}
		+ IFVD~\cite{54} & 65.96 $(\uparrow0.79)$ &  &   \hspace*{1cm}\\ 
   \hspace*{1cm}
		+ CWD~\cite{31} & 67.74 $(\uparrow2.57)$ &  &   \hspace*{1cm}\\
   \hspace*{1cm}
		+ CIRKD~\cite{29} & 68.18 $(\uparrow3.01)$ &  &   \hspace*{1cm}\\
   \hspace*{1cm}
		+ RDD (Ours) & $\textbf{69.78 (\bm{$\uparrow\mkern-5mu$}4.61)}$ &  &   \hspace*{1cm}\\
        \hdashline[2pt/1pt]
         \hspace*{1cm}
		S: DeepLabV3-MBV2 & 73.12 & \multirow{6}{*}{128.9G} &  \multirow{6}{*}{3.2M} \hspace*{1cm}\\
   \hspace*{1cm}
		+ SKD~\cite{28} & 73.82 $(\uparrow0.70)$ & &   \hspace*{1cm}\\
   \hspace*{1cm}
		+ IFVD~\cite{54} & 73.50 $(\uparrow0.38)$ & &   \hspace*{1cm}\\
   \hspace*{1cm}
		+ CWD~\cite{31} & 74.66 $(\uparrow1.54)$ & &   \hspace*{1cm}\\ 
   \hspace*{1cm}
		+ CIRKD~\cite{29} & 75.42 $(\uparrow2.30)$ & &   \hspace*{1cm}\\
   \hspace*{1cm}
		+ RDD (Ours) & $\textbf{76.38 (\bm{$\uparrow\mkern-5mu$}3.26)}$ & &   \hspace*{1cm}\\ 
        \hdashline[2pt/1pt]
         \hspace*{1cm}
		S: PSPNet-Res18 & 72.55 & \multirow{6}{*}{507.4G} & \multirow{6}{*}{12.9M}  \hspace*{1cm}\\
   \hspace*{1cm}
		+ SKD~\cite{28} & 73.29 $(\uparrow0.74)$ & &   \hspace*{1cm}\\
   \hspace*{1cm}
		+ IFVD~\cite{54} & 73.71 $(\uparrow1.16)$ & &   \hspace*{1cm}\\ 
   \hspace*{1cm}
		+ CWD~\cite{31} & 74.36 $(\uparrow1.81)$ & &   \hspace*{1cm}\\ 
   \hspace*{1cm}
		+ CIRKD~\cite{29} & 74.73 $(\uparrow2.18)$ &  &   \hspace*{1cm}\\ 
   \hspace*{1cm}
		+ RDD (Ours) & $\textbf{75.88 (\bm{$\uparrow\mkern-5mu$}3.33)}$ &  &    \hspace*{1cm}\\ 
\bottomrule
\end{tabular*}
\end{table}

\begin{inparaenum}[1)]
  \item \textbf{Results on Cityscapes.} 
To validate the performance of mIoU, we evaluate RDD on the Cityscapes~\cite{75} dataset.
We chose the following non-KD semantic segmentation models as student networks: DeepLabV3~\cite{17} based on ResNet-18 (Res18) backbone, DeepLabV3~\cite{17} based on MobileNetV2~\cite{80} backbone, and PSPNet~\cite{17} based on ResNet-18 (Res18) backbone. 
The experimental results are shown in Tab.~\ref{table:results_cityscapes}. 
All structured KD methods are observed to improve the student network's segmentation performance compared to training without knowledge distillation.
Our proposed RDD achieves optimal performance across all four student networks, demonstrating its robustness to variations in student network architecture.
Furthermore, for networks without ImageNet pre-training, RDD increases mIoU by 4.61\%.
This is mainly due to its progressive distillation design from easy to hard, which is more friendly to student networks with minimal knowledge.
Fig.~\ref{Fig:fig6} shows the qualitative segmentation results on the verification set using the DeepLabV3-ResNet18 network.
The validity of our proposed approach is intuitively demonstrated, and the semantic labels produced by RDD are more consistent with the ground truth.

Like other semantic segmentation methods based on knowledge distillation, RDD does not modify the model architecture. Therefore, the parameters (Params) of the model and the floating point operations (FLOPs) resulting from inference remain consistent with the underlying backbone network. Specifically, for all methods where the student model is DeepLabV3-Res18, the FLOPs are 572G and the Params are 13.6M. Similarly, for all methods where the student model is PSPNet-Res18, the FLOPs are 507.4G, and the Params are 12.9M.
Additionally, we verify the performance of our method on lightweight networks, such as using MobileNetV2 as the student network. As can be seen from the table, although the model based on MobileNetV2 has fewer parameters and fewer calculations (128.9G FLOPs, 3.2M Params), its mIoU reaches 76.38\%. It is even 0.5\% mIoU higher than the model based on PSPNet-Res18 (507.4G FLOPs,  12.9M Params) with larger parameters and calculations. This shows that our method is also suitable for some specially designed lightweight models, such as SqueezeNet~\cite{82} series, MobileNet~\cite{80} series, and ShuffleNet~\cite{83,84} series. RDD can achieve high segmentation accuracy while using as few computing resources and parameters as possible, thereby better deploying in resource-constrained environments such as embedded and mobile devices.

In Tab.~\ref{table:results_cityscapes}, we list several related non-KD semantic segmentation methods. On the one hand, lightweight models like ENet~\cite{22}, ESPNet~\cite{25}, and ICNet~\cite{23} require very little FLOPs. They are suitable for deployment on lightweight mobile devices but have relatively low mIoU. On the other hand, models like FCN~\cite{85}, RefineNet~\cite{86}, and OCRNet~\cite{87} have higher mIoU but are very computationally intensive and difficult to deploy on resource-constrained devices.
RDD methods achieve a certain balance between computational efficiency and segmentation performance compared to non-KD methods. This makes the RDD an effective semantic segmentation method, more suitable for resource-constrained devices.

\begin{figure}[!t]
\setlength{\abovecaptionskip}{0.1cm}

\centering
\includegraphics[width=1.0\textwidth]{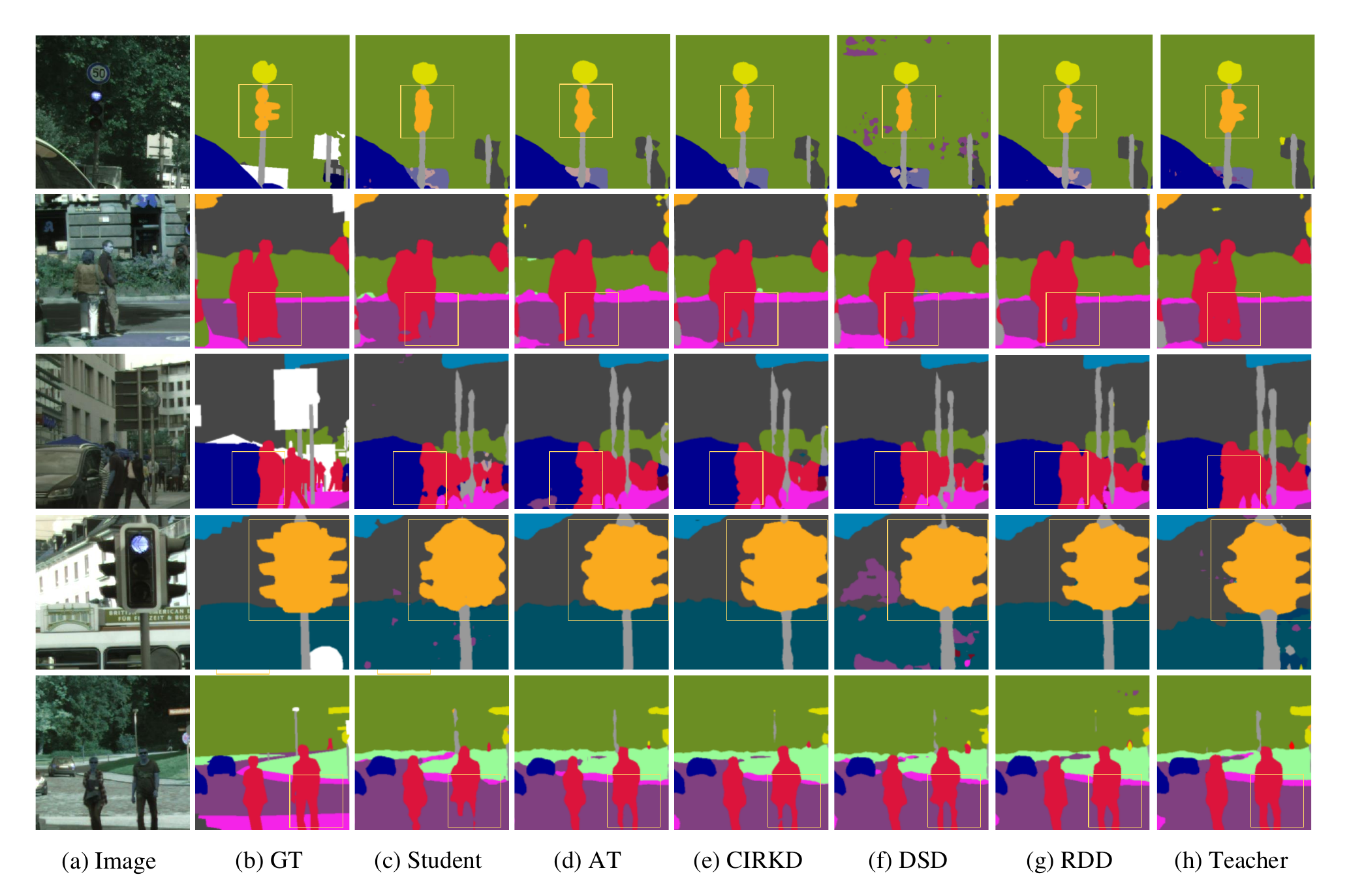}
\caption{Qualitative segmentation results on the validation set of Cityscapes using DeepLabV3-ResNet18 as student network and DeepLabV3-ResNet101 as teacher network: (a) Input image. (b) ground truth. (c) Results of original student network without KD. (d) Results of AT~\cite{52}. (e) Results of CIRKD~\cite{29}. (f) Results of DSD~\cite{74}. (g) Results of the proposed RDD. (h) Results of teacher network.
}
\label{Fig:fig6}
\end{figure}

\begin{table}[!h]
\footnotesize
\caption{Performance comparison with SOTA KD methods over various student segmentation networks on Pascal VOC 2012. FLOPs are measured based on the test size of 512 × 512.}
\label{table:results_voc2012}
\begin{tabular*}{1.0\textwidth}{@{\extracolsep\fill}>{\centering\hspace{1em}}c>{\centering\hspace{5em}}c>{\centering\hspace{3em}}cc}
\toprule[1.5pt]
  \hspace*{1cm}
		    	Method & mIoU $(\%)$ & FLOPs(G) & Params(M) \hspace*{1cm}\\ 
                \midrule[0.8pt]
                \hspace*{1cm}
		    	T: DeepLabV3-Res101 & 77.67 & 1294.6G & 61.1M \hspace*{1cm}\\ 
      
                \midrule[0.8pt]
                 \hspace*{1cm}
		    	S: DeepLabV3-Res18 & 73.21 & \multirow{6}{*}{305.0G} & \multirow{6}{*}{13.6M} \hspace*{1cm}\\ 
       \hspace*{1cm}
		    	+ SKD~\cite{28} & 73.51 $(\uparrow0.30)$ & &  \hspace*{1cm}\\ 
       \hspace*{1cm}
		    	+ IFVD~\cite{54} & 73.85 $(\uparrow0.64)$ & &  \hspace*{1cm}\\ 
       \hspace*{1cm}
		    	+ CWD~\cite{31} & 74.02 $(\uparrow0.81)$ & &  \hspace*{1cm}\\ 
       \hspace*{1cm}
		    	+ CIRKD~\cite{29} & 74.50 $(\uparrow1.29)$ & &  \hspace*{1cm}\\ 
       \hspace*{1cm}
		    	+ RDD (Ours) & $\textbf{74.88 (\bm{$\uparrow\mkern-5mu$}1.67)}$ & &  \hspace*{1cm}\\ 
                \hdashline[2pt/1pt]
                \hspace*{1cm}
		    	S: PSPNet-Res18 & 73.33 & \multirow{6}{*}{260.0G} & \multirow{6}{*}{12.9M} \hspace*{1cm}\\
       \hspace*{1cm}
		    	+ SKD~\cite{28} & 74.07 $(\uparrow0.74)$ & &  \hspace*{1cm}\\
       \hspace*{1cm}
		    	+ IFVD~\cite{54} & 73.54 $(\uparrow0.21)$ & &  \hspace*{1cm}\\ 
       \hspace*{1cm}
		    	+ CWD~\cite{31} & 73.99 $(\uparrow0.66)$ & &  \hspace*{1cm}\\ 
       \hspace*{1cm}
		    	+ CIRKD~\cite{29} & 74.78 $(\uparrow1.45)$ &  &  \hspace*{1cm}\\ 
       \hspace*{1cm}
		    	+ RDD (Ours) & $\textbf{74.82 (\bm{$\uparrow\mkern-5mu$}1.49)}$ &  &   \hspace*{1cm}\\ 
\bottomrule
\end{tabular*}
\end{table}

\begin{table}[!h]
\footnotesize
\caption{Performance comparison with SOTA KD methods over various student segmentation networks on ADE20k. FLOPs are measured based on the test size of 512 × 512.}
\label{table:results_ade20k}
\begin{tabular*}{1.0\textwidth}{@{\extracolsep\fill}>{\centering\hspace{1em}}c>{\centering\hspace{5em}}c>{\centering\hspace{3em}}cc}
\toprule[1.5pt]
  \hspace*{1cm}
		    	Method & mIoU $(\%)$ & FLOPs (G) & Params (M) \hspace*{1cm}\\ 
                \midrule[0.8pt]
                \hspace*{1cm}
		    	T: DeepLabV3-Res101 & 42.70 & 1294.6G & 61.1M \hspace*{1cm}\\ \midrule[0.8pt]
       \hspace*{1cm}
		    	S: DeepLabV3-Res18 & 36.52 & \multirow{6}{*}{305.0G} & \multirow{6}{*}{13.6M}  \hspace*{1cm}\\ 
       \hspace*{1cm}
                    + KD~\cite{30} & 36.63 $(\uparrow0.11)$ &  &  \hspace*{1cm}\\
                    \hspace*{1cm}
                    + AT~\cite{52} & 37.79 $(\uparrow1.27)$ &  &   \hspace*{1cm}\\
                    \hspace*{1cm}
		    	+ DSD~\cite{74} & 37.37 $(\uparrow0.85)$ &  &   \hspace*{1cm}\\ 
       \hspace*{1cm}
		    	+ RDD (Ours) & $\textbf{38.04 (\bm{$\uparrow\mkern-5mu$}1.52)}$ & &  \hspace*{1cm}\\ 
\bottomrule
\end{tabular*}
\end{table}

\begin{table}[!h]
\footnotesize
\caption{Performance comparison with SOTA KD methods over various student segmentation networks on CamVid. FLOPs are measured based on the test size of 360 × 480.}
\label{table:results_camvid}
\begin{tabular*}{1.0\textwidth}{@{\extracolsep\fill}>{\centering\hspace{1em}}c>{\centering\hspace{5em}}c>{\centering\hspace{3em}}cc}
\toprule[1.5pt]
  {\hspace{1cm}}
		    	Method & mIoU $(\%)$ & FLOPs (G) & Params (M) \hspace*{1cm}\\ \midrule[0.8pt]
       {\hspace{1cm}}
		    	T: DeepLabV3-Res101 & 69.84 & 280.2G & 61.1M \hspace*{1cm}\\ \midrule[0.8pt]
       \hspace*{1cm}
		    	S: DeepLabV3-Res18 & 66.92 & \multirow{6}{*}{61.0G} & \multirow{6}{*}{13.6M} \hspace*{1cm}\\ 
       {\hspace{1cm}}
		    	+ SKD~\cite{28} & 67.46 $(\uparrow0.54)$ & &  \hspace*{1cm}\\ 
       {\hspace{1cm}}
		    	+ IFVD~\cite{52} & 67.28 $(\uparrow0.36)$ & &  \hspace*{1cm}\\ 
       {\hspace{1cm}}
		    	+ CWD~\cite{31} & 67.71 $(\uparrow0.79)$ & &  \hspace*{1cm}\\ 
       {\hspace{1cm}}
		    	+ CIRKD~\cite{29} & 68.21 $(\uparrow1.29)$ & &  \hspace*{1cm}\\ 
       {\hspace{1cm}}
		    	+ RDD (Ours) & $\textbf{68.55 (\bm{$\uparrow\mkern-5mu$}1.63)}$  & & \hspace*{1cm}\\ 

                \hdashline[2pt/1pt]
        \hspace*{1cm}
		    	S: PSPNet-Res18 & 66.73 & \multirow{6}{*}{45.6G} & \multirow{6}{*}{12.9M} \hspace*{1cm}\\ 
       {\hspace{1cm}}
		    	+ SKD~\cite{28} & 67.83 $(\uparrow1.10)$ & &  \hspace*{1cm}\\
       {\hspace{1cm}}
		    	+ IFVD~\cite{52} & 67.61 $(\uparrow0.88)$ & &  \hspace*{1cm}\\ 
       {\hspace{1cm}}
		    	+ CWD~\cite{31} & 67.92 $(\uparrow1.19)$ & &  \hspace*{1cm}\\ 
       {\hspace{1cm}}
		    	+ CIRKD~\cite{29} & 68.65 $(\uparrow1.92)$ &  &  \hspace*{1cm}\\ 
       {\hspace{1cm}}
		    	+ RDD (Ours) & $\textbf{68.77 (\bm{$\uparrow\mkern-5mu$}2.04)}$ &  &   \hspace*{1cm}\\
\bottomrule
\end{tabular*}
\end{table}
\item \textbf{Results on Pascal VOC 2012.} 
We also evaluate RDD on Pascal VOC 2012~\cite{77}, a representative visual object segmentation dataset. We use the DeepLabV3~\cite{17} with the backbone of ResNet-18 (Res18) and PSPNet~\cite{20} with the backbone of ResNet-18 (Res18) as two student networks.
As shown in Tab.~\ref{table:results_voc2012}, RDD outperforms other KD methods based on semantic segmentation.
Compared to the original student networks, RDD improves mIoU by 1.67
Specifically, for DeepLabV3 with ResNet-18 backbone, mIoU increased from 73.21\% to 74.88\%. For PSPNet with ResNet-18 backbone, mIoU improved from 73.33\% to 74.83\%.

 \item \textbf{Results on ADE20k.} 
Tab.~\ref{table:results_ade20k} compares the performance of RDD with state-of-the-art distillation methods on the ADE20k dataset.
We use the DeepLabV3~\cite{17} with ResNet-18 (Res18) backbone as the student network.
RDD achieves the best performance compared to standard KD~\cite{30}, AT~\cite{52}, and DSD~\cite{74}.
For DeepLabV3 with ResNet-18 backbone, mIoU on the validation set increases from 36.52\% to 38.04\%. 
Our method based on the same network also outperforms CIRKD's result of 37.07\%.
The experimental results demonstrate the effectiveness and generality of our approach, and RDD achieves consistent performance improvement in different segmentation networks and further narrows the performance gap with teacher networks.

 \item \textbf{Results on CamVid.} 
We evaluate various distillation methods on CamVid (a simple small scene understanding dataset) dataset shown in Tab.~\ref{table:results_camvid}. RDD consistently achieves optimal performance.
We use the DeepLabV3~\cite{17} with  ResNet-18 (Res18) backbone and the PSPNet~\cite{20} with  ResNet-18 (Res18) backbone as two student networks.
Due to the simplicity of the dataset, the compact student network can obtain segmentation performance close to the cumbersome teacher network. However, RDD can still improve the upper bound of the performance of the student network. 
Compared to the original student networks, our method improves the mIoU of the two student networks by 1.63\% and 2.04\%, respectively.
Specifically, for DeepLabV3 with ResNet-18 backbone, mIoU increased from 66.92\% to 68.55\%. For PSPNet with ResNet-18 backbone, mIoU improved from 66.73\% to 68.77\%.
\end{inparaenum}

\subsection{Ablation study}\label{subsec4.4}
\subsubsection{Ablation experiment for the two-stage switch}\label{subsec4.4.1}
We conduct ablation experiments on the Cityscape dataset to evaluate the effectiveness of the two-stage switch in RDD.
For these experiments, we employ the segmentation framework DeepLabV3~\cite{17} with ResNet-101 (Res101) backbone~\cite{71} as the powerful teacher network and DeepLabV3~\cite{17} with ResNet-18 (Res18) backbone as the student network.
The ablation experimental results are summarized in Tab.~\ref{table:ablation_study}.
RDD consists of two stages: TFE-RDD and TSE-RDD.
The experimental group (a) indicates that only TFE-RDD is involved throughout the whole training process, and the experimental group (b) indicates that only TSE-RDD is involved in the whole training process.
Experimental groups (c)-(e) investigate the role played by TFE-RDD and the proportion's effect of training phase iterations accounted for by TE-REDD on the performance of the student network.

We adopted warm-up training in the early learning stage.
The teacher network provides supervision information to guide the student network to learn relatively simple pixels and achieve rapid convergence. This stage of training should be short to avoid a warm-up period that is too long and cannot give full play to the potential of the student model itself. To determine the optimal percentage parameter $p$ of training iterations, we tried four different percentages: 0\%, 10\%, 20\%, 30\%. By evaluating the performance of the student network at different scales, we aim to find the most suitable parameter $p$ to achieve a balance between fast learning and giving full play to the potential of the student network.
The group (c) indicates that the percentage of TFE-RDD stage iterations is 0\%, the same as the experimental group (b).

The experimental results for groups (a) and (b) indicate that the TSE-RDD stage significantly affects the network performance. Without TSE-RDD, the performance is almost the same as the baseline, and the network only has an improvement of 0.22\% mIoU.
According to the analysis, without the TSE-RDD stage involved in training, the network consistently tends to focus primarily on learning simple pixels. It keeps the learning preference for difficult pixels at a low level.
This phenomenon is similar to the comfort zone in human education, where too much simple training does not easily raise the learner's upper performance bound. 
Experimental groups (d), (e), and (f) evaluated the contribution of the TFE-RDD stage. 
The results suggest that the TFE-RDD stage is the icing on the cake, and the performance is further improved.
Through the analysis, we find that in the early learning stage, the student network has not yet converged, and the judgment of difficult pixels is not stabilizable, so the direct application of TSE-RDD does not work well.
The experimental results show that applying TFE-RDD during the initial 10\% of training iterations and TSE-RDD in the remaining iterations yields the best KD performance.
\begin{table}[!t]
\footnotesize
\caption{The effect of components in the proposed method.}
\label{table:ablation_study}
\begin{tabular*}{\textwidth}{@{\extracolsep\fill}ccccc}
\toprule[1.5pt]
   &  \multicolumn{2}{c}{Distillation} & Training & \\ \midrule[0.8pt] 
         {\hspace{1cm}}
		    	Method  &  TFE-RDD & TSE-RDD & Training Iterations $p$ & mIoU $(\%)$ \hspace*{1cm}\\ \midrule[0.8pt] 

		    	\multicolumn{4}{l}{T: DeepLabV3-Res101} & 78.07 \hspace*{1cm}\\ 
		    	\multicolumn{4}{l}{S: DeepLabV3-Res18}  & 74.21 \hspace*{1cm}\\ \midrule[0.8pt]
       {\hspace{1cm}}
		    	(a) & $\checkmark$ & & & 74.43 $(\uparrow0.22)$ \hspace*{1cm}\\ 
       {\hspace{1cm}}
		    	(b) & & $\checkmark$ & & 76.34 $(\uparrow2.13)$ \hspace*{1cm}\\ \midrule[0.8pt]
       {\hspace{1cm}}
		    	(c) & $\checkmark$ & $\checkmark$ & 0\%  & 76.34 $(\uparrow2.13)$ \hspace*{1cm}\\
       {\hspace{1cm}}
		    	(d) & $\checkmark$ & $\checkmark$ & 10\%  & $\textbf{77.18 (\bm{$\uparrow\mkern-5mu$}2.97)}$ \hspace*{1cm}\\
       {\hspace{1cm}}
		    	(e) & $\checkmark$ & $\checkmark$ & 20\%  &  76.77 $(\uparrow2.56)$ \hspace*{1cm}\\
        {\hspace{1cm}}
		    	(f) & $\checkmark$ & $\checkmark$ & 30\%  &  76.58 $(\uparrow2.37)$ \hspace*{1cm}\\
\bottomrule
\end{tabular*}
\end{table}

\subsubsection{Ablation experiment of TSE-RDD mode}\label{subsec4.4.2}
TSE-RDD's combination of the teacher and student networks can be expressed as different bitwise operations, including \rm{XOR}, \rm{AND}, and \rm{OR} operations.
In our TSE-RDD stage, the relative difficulty $\mathit{RD}_\mathit{TS}$ obtained by the \rm{XOR} operation between the teacher and the student networks is expressed as $\mathit{RD_{TSE}} = (f(x \mid \theta_{s}) \leq t) \oplus (f(x \mid \theta_{t}) \leq t)$.
The strategy focuses on selecting pixels where the teacher network's predictions are above threshold $t$ while the student network's predictions are below threshold $t$.
It encourages the student network to learn the difficult pixels that the teacher network has mastered and the student network has not yet mastered but is capable of learning.
Similarly, when using an \rm{AND} operation between the teacher and student networks, we obtain $\mathit{RD_{TSE}} = (f(x \mid \theta_{s}) \leq t) \land (f(x \mid \theta_{t}) \leq t)$.
This strategy selects pixels where the prediction confidences of both the teacher and student networks are below threshold $t$. 
It encourages the student network to learn parts of pixels that are difficult for even the teacher network to master.
Lastly, we consider an \rm{OR} operation between them:  
$\mathit{RD_{TSE}} = (f(x \mid \theta_{s}) \leq t) \lor (f(x \mid \theta_{t}) \leq t)$.
This strategy combines results obtained from both \rm{XOR} and \rm{AND} operations. 
It encourages the student network to learn difficult pixels that promise to be learned well and difficult pixels that even the teacher network struggles to master.

\begin{table}[!t]
\footnotesize
\caption{The effect of combination mode of teacher and student difficulty maps in the TSE-RDD stage.}
\label{table:ablation_study_mode}
\begin{tabular*}{\textwidth}{@{\extracolsep\fill}ccccc}
\toprule[1.5pt]
  &  \multicolumn{2}{c}{Distillation} & Training & \\                
                    \midrule[0.8pt]
                    {\hspace{1cm}}
		    	Method  &  TFE-RDD & TSE-RDD & Training Iterations $p$ & mIoU $(\%)$  \hspace*{1cm}\\ \midrule[0.8pt]
		    	\multicolumn{4}{l}{T: DeepLabV3-Res101} & 78.07 \hspace*{1cm}\\ 
		    	\multicolumn{4}{l}{S: DeepLabV3-Res18}  & 74.21 \hspace*{1cm}\\ \midrule[0.8pt]
                    {\hspace{1cm}}
                    AND     & $\checkmark$ & $\checkmark$ & 10\%  & 70.01 \hspace*{1cm}\\
                    {\hspace{1cm}}
		    	OR      & $\checkmark$ & $\checkmark$ & 10\%  & 75.32 \hspace*{1cm}\\ 
                    {\hspace{1cm}}
		    	XOR     & $\checkmark$ & $\checkmark$ & 10\%  & \textbf{77.18} \hspace*{1cm}\\
\bottomrule
\end{tabular*}
\end{table}

\begin{table}[!t]
\footnotesize
\caption{The effect of threshold $t$ in the TSE-RDD stage.}
\label{table:threshold t}
\begin{tabular*}{\textwidth}{@{\extracolsep\fill}cccccc}
\toprule[1.5pt]
  \hspace*{1cm}
		        threshold $t$  &  0.60 & 0.65 & 0.70  &0.75 &0.80   \hspace*{1cm}\\ \midrule[0.8pt]
           \hspace*{1cm}
                    mIoU$(\%)$     & 76.47 & 76.88 & \textbf{76.95}  & 76.66 & 76.32
		    	 \hspace*{1cm}\\
\bottomrule
\end{tabular*}
\end{table}

We conduct ablation experiments on the influence of these three bitwise operation modes on our TSE-RDD stage. 
The experimental results are shown in Tab.~\ref{table:ablation_study_mode}. 
The \rm{OR} mode's mIoU decreased by 1.86\% compared to the \rm{XOR} mode. 
Analysis reveals that compared to \rm{XOR} mode, \rm{OR} mode also introduces pixels that are difficult even for the teacher network to master, and these pixels can be difficult pixels with noise annotations or semantic ambiguities. These challenging pixels can negatively impact network performance.
The \rm{AND} mode achieved mIoU of 70.01\%, which is 4.20\% lower than the student network without the knowledge distillation method. 
Analysis reveals that \rm{AND} mode introduces too many difficult pixels for the student network and only encourages students to learn difficult pixels that even the teacher network struggles to master. The \rm{AND} mode's experimental results indicate that using excessively difficult pixels only in the later stages of training would severely hinder student network learning.
In contrast, the \rm{OR} mode not only discards the simple pixels that the student network has already mastered and does not need to learn but also separates difficult pixels from those that may be noisy annotations or have semantic ambiguities. It selectively learns the most valuable difficult pixels that the student network is expected to master. As a result, the \rm{OR} mode achieves the best performance.

\subsubsection{Ablation experiments with threshold in TSE-RDD}\label{subsec4.4.3}
We investigate the effect of threshold $t$ on determining the number of difficult pixels to retain in the TSE-RDD stage.
In the MMSegmentation code base~\cite{88}, the confidence threshold used to filter difficult samples is set to 0.7. To further investigate the impact of threshold $t$, we performed a set of ablation experiments.
In the experiments, we use the DeepLabV3~\cite{17} with ResNet-101 (Res101) backbone~\cite{71} as the  teacher network and DeepLabV3~\cite{17} with ResNet-18 (Res18) backbone as the student network.
The experimental results are presented in Tab.~\ref{table:threshold t} and averaged over three independent runs.
It can be observed that an appropriate threshold $t$ is effective in providing an optimal number of difficult pixels, leading to improved learning for the student network. However, setting a threshold $t$ that is too large or too small has a negative impact on the student network's learning.
Analysis reveals that setting threshold $t$ too small would discard too many relatively difficult pixels, thus failing to enhance the student network's upper performance bound. Conversely, when threshold $t$ is set excessively large, many simple samples will be retained. It makes it difficult to play out the learning that TSE-RDD would have done by utilizing difficult pixels.
The experimental results show that the RDD method performs best when $t$ = 0.70.


\subsection{Integrating with other approaches and training time}\label{subsec4.5}

\begin{figure}[!h]

    \centering
    \includegraphics[width=1.0\textwidth]{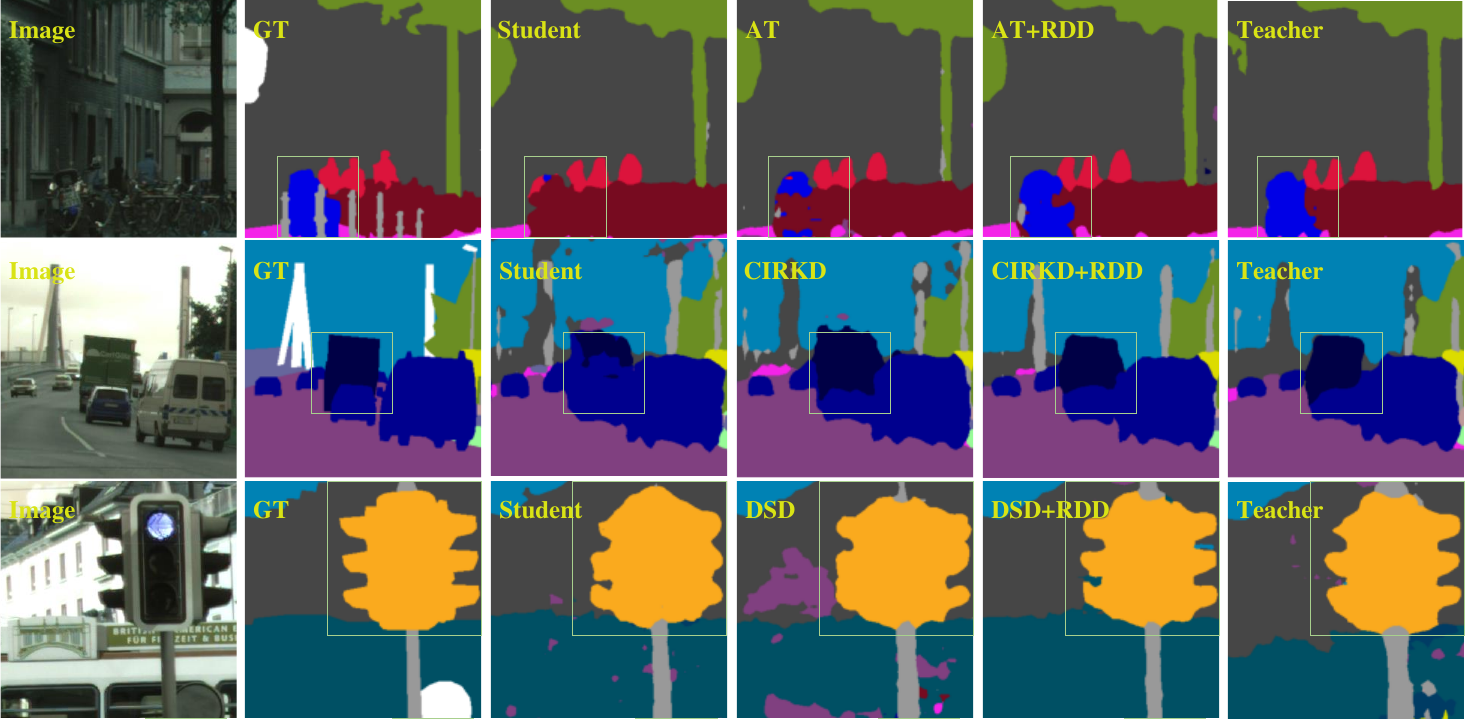}
    \caption{Predictions of KD methods with and without integrating RDD.}
    \label{Fig:Integrate}

\end{figure}

We evaluate integrating RDD's impact into existing KD approaches, including AT~\cite{52}, DSD~\cite{74}, CIRKD~\cite{29}, and KD~\cite{30}.
The KD group only uses the spatial distillations shown in Eq.~\ref{eq:loss_kd}.
In these methods, we use DeepLabV3-Res18 as the student model. When integrating with other methods, we did not change any hyperparameters under the original method's experimental settings to explore our method's effectiveness in the simplest integration mode.
The experimental results are presented in Tab.~\ref{table:results_integration}.
In the four baseline approaches, RDD effectively improves the performance of all approaches, further narrowing the performance gap between the student and teacher networks. 
After integrating RDD, mIoU for each approach shows improvement. 
Specifically, the AT method's mIoU after integrating RDD increases by 0.9\%. 
DSD method's mIoU after integrating RDD increases by 1.3\%. 
CIRKD method's mIoU after integrating RDD increases by 0.31\%. 
For the KD method, the mIoU increases by 2.36\% after integration.
Fig.~\ref{Fig:Integrate} compares prediction maps for methods with and without RDD integration. The visualization results of the methods integrated with RDD generally perform better on complex regions such as edges and obscured objects.

Furthermore, we access the training time before and after integrating various KD methods with RDD. Table \ref{table:timecompare} displays the experimental results comparing training times.
As can be seen from the table, among the many KD methods, RDD requires the shortest training time, which is only 4 hours and 20 minutes. In comparison, the training time of AT, CIRKD, and DSD methods is longer than RDD. It is worth noting that the CIRKD method employs a memory bank for contrastive learning and needs to optimize five loss terms, which results in its training time of up to 7 hours and 38 minutes, almost twice as long as other methods. This also verifies the problem that adding multiple optimization objectives will increase training difficulty and time.
This comparison concludes that integrating RDD does not incur excessive computational overhead or training time.
It only incurs an average increase of less than 10\% in training time across different methods.

\begin{table}[!t]
\footnotesize
\caption{Integrating with other KD approaches. "+" denotes implementing the corresponding schemes.}
\label{table:results_integration}
\begin{tabular*}{1.0\textwidth}{@{\extracolsep\fill}cc>{\centering\hspace{2em}}cc}
\toprule[1.5pt]
  \hspace*{1cm}
		    	Method & mIoU $(\%)$ & FLOPs (G) & Params (M)  \hspace*{1cm}\\ \midrule[0.8pt]
        \hspace*{1cm}
		    	T: DeepLabV3-Res101 & 78.07 & 2371G & 61.1M  \hspace*{1cm}\\ \midrule[0.8pt]
        \hspace*{1cm}
		    	S: DeepLabV3-Res18 & 74.21 & \multirow{1}{*}{572.0G} & \multirow{1}{*}{13.6M}  \hspace*{1cm}\\ \midrule[0.8pt]
        \hspace*{1cm}
		    	+ AT~\cite{52} & 76.21 & \multirow{2}{*}{572.0G} & \multirow{2}{*}{13.6M}  \hspace*{1cm}\\
        \hspace*{1cm}
		    	+ AT~\cite{52} + RDD (Ours) & \textbf{77.11} $(\uparrow0.90)$ & &   \hspace*{1cm}\\ 
                 \hdashline[2pt/1pt]
                  \hspace*{1cm}
		    	+ DSD~\cite{74} & 74.42 & \multirow{2}{*}{572.0G} & \multirow{2}{*}{13.6M}  \hspace*{1cm}\\ 
		    	+ DSD~\cite{74} + RDD (Ours) & \textbf{75.72} $(\uparrow1.30)$ & &   \hspace*{1cm}\\ 
                 \hdashline[2pt/1pt]
                  \hspace*{1cm}
		    	+ CIRKD~\cite{29} & 76.38 & \multirow{2}{*}{572.0G} & \multirow{2}{*}{13.6M}  \hspace*{1cm}\\
        \hspace*{1cm}
		    	+ CIRKD~\cite{29} + RDD (Ours) & \textbf{76.69} $(\uparrow0.31)$ & &   \hspace*{1cm}\\ 
                 \hdashline[2pt/1pt]
                  \hspace*{1cm}
		    	+ KD~\cite{30} & 74.82 & \multirow{2}{*}{572.0G} & \multirow{2}{*}{13.6M}  \hspace*{1cm}\\ 
        \hspace*{1cm}
		    	+ KD~\cite{30} + RDD (Ours) & \textbf{77.18} $(\uparrow2.36)$ & &   \hspace*{1cm}\\
\bottomrule
\end{tabular*}
\end{table}

\begin{table}[!t]
\footnotesize
\caption{Comparison of training time for experiments with and without RDD integration.}
\label{table:timecompare}
\begin{tabular*}{\textwidth}{@{\extracolsep\fill}cccc}
\toprule[1.5pt]
   \hspace*{1cm}
		    	Method   & Cost  & Method with RDD   & Cost   \hspace*{1cm}\\
		    \midrule[0.8pt]
       \hspace*{1cm}
                Baseline & 4 $h$ 05 $m$ & Baseline with RDD & 4 $h$ 20 $m$  \hspace*{1cm}\\
                 \hspace*{1cm}
                AT~\cite{52}       & 4 $h$ 30 $m$ & AT~\cite{50} with RDD       & 4 $h$ 52 $m$  \hspace*{1cm}\\
                 \hspace*{1cm}
                CIRKD~\cite{29}    & 7 $h$ 38 $m$ & CIRKD~\cite{29} with RDD    & 8 $h$ 02 $m$  \hspace*{1cm}\\
                 \hspace*{1cm}
                DSD~\cite{74}      & 4 $h$ 32 $m$ & DSD~\cite{74} with RDD      & 4 $h$ 56 $m$  \hspace*{1cm}\\
\bottomrule
\end{tabular*}
\end{table}

\section{Conclusion and future work}\label{sec5}
We propose Relative Difficulty Distillation (RDD) for Semantic Segmentation. 
Unlike previous feature- and response-based distillation methods, we leverage the relative difficulty of samples as dark knowledge transfers from teacher to student and adjust the learning procedure based on it.
We propose TFE-RDD and TSE-RDD for use at different stages of student network training. 
By tending to learn simpler pixels in the early stage to achieve rapid convergence and focusing on valuable difficult pixels in the later stage to enhance the performance ceiling, the student network effectively improves its overall performance.
RDD is easy to integrate with existing distillation methods. Experimental results demonstrate RDD outperforms state-of-the-art knowledge distillation methods. 

In this work, we focus on the relative difficulty distillation of the semantic segmentation task. 
However, integrating RDD into other vision tasks, such as classification and detection, is also feasible.
In future research, we will further explore other potential effects of the relative difficulty of knowledge distillation and explore adaptability in tasks beyond semantic segmentation.

\Acknowledgements{
This work was partly supported by NSFC (Grant Nos. 62272229, 62076124, 62222605), the National Key R$\&$D Program of China (2020AAA0107000), the Natural Science Foundation of Jiangsu Province (Grant Nos. BK20222012, BK20211517), and Shenzhen Science and Technology Program JCYJ20230807142001004. The authors would like to thank all the anonymous reviewers for their constructive comments.}





\end{document}